\documentstyle[latex-acl,epsf]{article}

\newcommand{\tuple}[1]{\mbox{$\langle$#1$\rangle$}}

\title{Tricolor DAGs for Machine Translation}
\author{Koichi Takeda \\ \\
    IBM Research, Tokyo Research Laboratory \\
    1623-14 Shimotsuruma, Yamato, Kanagawa 242, Japan \\
    Phone: 81-462-73-4569, 81-462-73-7413 (FAX) \\
    takeda@trl.vnet.ibm.com
}

\begin{document}
\maketitle

\begin{abstract}
Machine translation (MT) has recently been formulated
in terms of constraint-based
knowledge representation and unification theories, but it is becoming
more and more evident that it is not possible to design a practical MT
system without an adequate method of handling mismatches between semantic
representations in the
source and target languages. In this paper, we introduce the idea of
``information-based''
MT, which is considerably more flexible than interlingual MT or the
conventional transfer-based MT.
\end{abstract}

\section{Introduction}

With the intensive exploration of contemporary theories on
unification grammars\cite{kaplb82,shie86,polls87} and feature
structures\cite{kaspr8606,smol88} in the
last decade, the old image of machine translation (MT) as
a brutal form of natural language processing has given way to that of
a process based on a uniform and reversible
architecture\cite{shie88,dymetman91,zajac91}.

The developers of MT systems based on the constraint-based formalism
found a serious problem in ``language mismatching,'' namely,
the difference between semantic representations in the source
and target languages.\footnote{For example, Yasuhara\cite{yasuhara93}
reported there was an overlap of
only 10\% between his group's English and Japanese
concept dictionaries, which covered 0.2 million concepts.}
Attempts to design a pure interlingual MT system were therefore
abandoned,\footnote{Even an MT system
with a controlled input language\cite{kant} does not claim to be a pure
interlingual system.} and the notion of
``semantic transfer''\cite{uchida88,takeu9207} came into
focus as a practical solution to the problem of handling the language
mismatching.
The constraint-based formalism\cite{emele92} seemed promising as
a formal definition of transfer, but pure constraints are too rigid to be
precisely imposed on target-language sentences.

Some researchers(e.g., Russell\cite{russell92})
introduced the concept of {\it defeasible reasoning}
in order to formalize what is missing
from a pure constraint-based approach, and control mechanisms for such
reasoning have also been proposed\cite{hobbs93,hasida91}. With this
additional mechanism, we can formulate the ``transfer'' process as a mapping
from a set of constraints into another set of mandatory and defeasible
constraints. This idea leads us further to the concept of
``information-based'' MT,
which means that, with an appropriate representation scheme, a source
sentence can be represented by a set of constraints that it implies and
that, given a target sentence, the set $C_{s}$ of constraints can be
divided into three disjoint subsets:
\begin{itemize}
\item The subset $C_{0}$ of constraints that is also implied by the
target sentence
\item The subset $C_{+}$ of constraints that is not implied by, but is
consistent with, the translated sentence
\item The subset $C_{-}$ of constraints that is
violated by the target sentence
\end{itemize}
The target sentence may also imply another set $C_{new}$ of constraints,
none of which is in $C_{s}$. That is, the set $C_{t}$ of
constraints implied by the target sentences is a union of $C_{0}$ and
$C_{new}$, while $C_{s} = C_{0} \cup C_{+} \cup C_{-}$. When $C_{s}
= C_{0} = C_{t}$, we have a {\it fully interlingual} translation of the
source sentence. If $C_{+} \neq \phi, C_{-} = \phi$, and $C_{new} = \phi$,
the target sentence is said to be {\it under-generated}, while it is said
to be {\it over-generated} when $C_{+} = \phi, C_{-} = \phi$, and
$C_{new} \neq \phi$.\footnote{The notions of {\it completeness} and
{\it coherence} in LFG\cite{kaplb82} have been employed by
Wedekind\cite{wede88} to avoid over- and under-generation.}
In either case, $C_{-}$ must be empty if a
consistent translation is required. Thus, the goal of machine translation
is to find an optimal pair of source and target sentences that
minimizes $C_{+}, C_{-}$, and $C_{new}$. Intuitively, $C_{0}$ corresponds to
essential information, and $C_{+}$ and $C_{new}$ can be viewed as
language-dependent supportive information. $C_{-}$ might be the
inconsistency between the assumptions of the source- and target-language
speakers.

In this paper, we introduce {\it tricolor DAGs} to represent the above
constraints, and discuss how tricolor DAGs are used for practical MT
systems. In particular, we give a generation algorithm that
incorporates the notion of semantic transfer by gradually approaching
the optimal target sentence through the use of tricolor DAGs, when a fully
interlingual translation fails.
Tricolor DAGs give a
{\it graph-algorithmic} interpretation of the constraints, and the
distinctions between the types of constraint mentioned above allow us to
adjust the margin between the current and optimal solution effectively.

\section{Tricolor DAGs}

A {\it tricolor DAG} (TDAG, for short)
is a rooted, directed, acyclic\footnote{Acyclicity
is not crucial to the results in this paper, but it significantly
simplifies the definition of the tricolor DAGs and semantic transfer.}
graph with a set of three
colors (red, yellow, and green) for nodes and directed arcs.
It is used to represent a feature structure of a source or target
sentence. Each node represents either an atomic value or a root of
a DAG, and each arc is labeled with a feature name. The only difference
between the familiar usage of DAGs in unification grammars and that of
TDAGs is that the color of a node or arc represents its degree of
importance:
\begin{enumerate}
\item Red shows that a node (arc) is essential.
\item Yellow shows that a node (arc) may be ignored,
but must not be violated.
\item Green shows that a node (arc) may be violated.
\end{enumerate}
For practical reasons, the above distinctions are interpreted as follows:
\begin{enumerate}
\item Red shows that a node (arc) is derived from lexicons
and grammatical constraints.
\item Yellow shows that a node (arc) may be inferred from a source or a
target sentence by using domain knowledge, common sense, and so on.
\item Green shows that a node (arc) is defeasibly inferred, specified
as a default, or heuristically specified.
\end{enumerate}
When all the nodes and arcs of TDAGs are red, TDAGs are basically the same
as the feature structures\footnote{We will only consider the semantic
portion of the feature structure although the theory of tricolor DAGS for
representing entire feature structures is an interesting topic.}
of grammar-based translation\cite{wede88,shiep90}.
A TDAG is {\it well-formed} iff the following conditions are satisfied:
\begin{enumerate}
\item The root is a red node.
\item Each red arc connects two red nodes.
\item Each red node is reachable from the root through the red
arcs and red nodes.
\item Each yellow node is reachable from the root through the arcs
and nodes that are red and/or yellow.
\item Each yellow arc connects red and/or yellow nodes.
\item No two arcs start from the same node, and have the
same feature name.
\end{enumerate}
Conditions 1 to 3 require that all the red nodes and red arcs
between them make a single, connected DAG. Condition 4 and 5 state that
a defeasible constraint must not be used to derive an imposed constraint.
In the rest of this paper, we will consider only well-formed TDAGs.
Furthermore, since
only the semantic portions of TDAGs are used for machine translation,
we will not discuss syntactic features.

The {\it subsumption} relationship among the TDAGs is defined as
the usual subsumption over DAGs, with the following extensions.
\begin{itemize}
\item A red node (arc) subsumes only a red node (arc).
\item A yellow node (arc) subsumes a red node (arc) and a yellow node (arc).
\item A green node (arc) subsumes a node (arc) with any color.
\end{itemize}
The {\it unification} of TDAGs is similarly defined. The colors of unified
nodes and arcs are specified as follows:
\begin{itemize}
\item Unification of a red node (arc) with another node (arc) makes a
red node (arc).
\item Unification of a yellow node (arc) with a yellow or green node
(arc) makes a yellow node (arc).
\item Unification of two green nodes (arcs) makes a green node (arc).
\end{itemize}
Since the green nodes and arcs represent defeasible constraints,
unification of a green node (either a root of a TDAG or an atomic node)
with a red or yellow node always succeeds, and results in a red or
yellow node. When two conflicting green nodes are to be unified, the
result is {\it indefinite}, or a single non-atomic green
node.\footnote{An alternative definition is that one green node
has precedence over the other\cite{russell92}. Practically, such a
conflicting unification should be postponed until no other possibility is
found.}

\begin{figure}
\centering{\epsfile{file=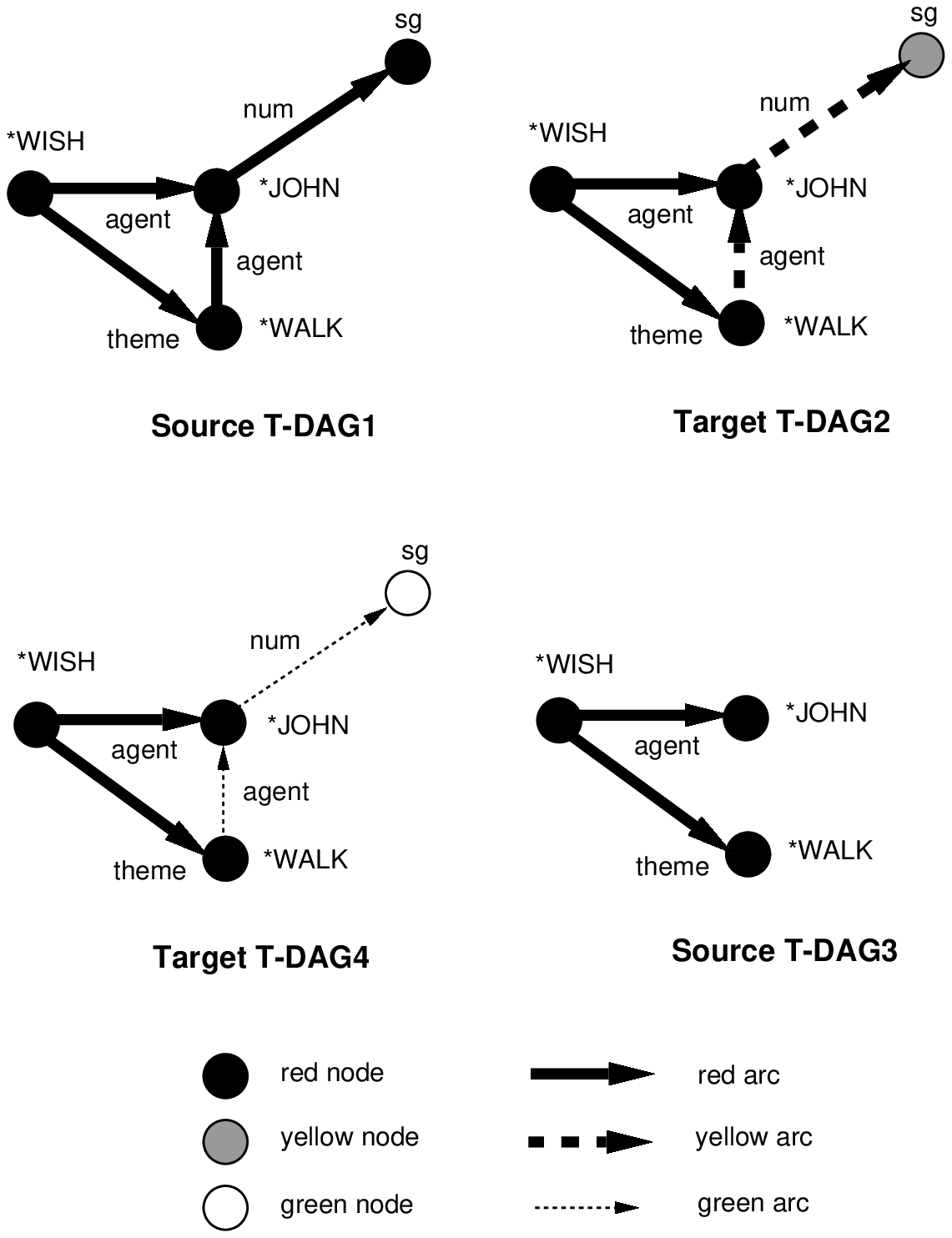,width=8cm,height=10cm}}
\caption{Sample TDAGs}
\label{tdag}
\end{figure}

Now, the problem is that a red node/arc in a {\it source TDAG} (the TDAG
for a source sentence) may not
always be a red node/arc in the {\it target TDAG} (the TDAG for a target
sentence).
For example, the
{\it functional control} of the verb ``wish'' in the English sentence
\begin{verbatim}
   John wished to walk
\end{verbatim}
may produce the $TDAG_{1}$ in Figure~\ref{tdag},
but the red arc corresponding
to the {\it agent} of the *WALK predicate
may not be preserved in a target $TDAG_{2}$.\footnote{For example,
the Japanese counterpart ``nozomu''
for the verb ``wish'' only takes a sentential complement, and no
functional control is observed.} This means that the target sentence alone
cannot convey the information that it is John who wished to walk,
even if this information can be understood from the context.
Hence the red arc is relaxed into a yellow one, and any target
TDAG must have an agent of *WALK that is consistent with *JOHN.
This relaxation will help the sentence generator in two ways. First,
it can prevent generation failure (or non-termination in the worst case).
Second, it retains important information for a choosing correct
translation of the verb ``walk''.\footnote{Whether or not the subject
of the verb is human is often crucial information
for making an appropriate choice between the verb's
two Japanese counterparts ``aruku'' and ``hokousuru''.}

Another example is the problem of identifying
{\it number} and {\it determiner}
in Japanese-to-English translation. This type of
information is rarely available from a syntactic representation of a
Japanese noun phrase, and a set of heuristic rules\cite{murata93}
is the only known basis for making a reasonable guess. Even if such
contextual
processing could be integrated into a logical inference system,
the obtained information should be defeasible, and hence should
be represented by green nodes and arcs in the TDAGs. Pronoun
resolution can be similarly represented by using green nodes
and arcs.

It is worth looking at the source and target TDAGs in the opposite
direction. From the Japanese sentence,
\begin{verbatim}
John   ha   aruku koto  wo  nozonda
John  +subj walk +nom  +obj wished
\end{verbatim}
we get the source $TDAG_{3}$ in Figure~\ref{tdag}, where functional
control and number information are missing. With the help of contextual
processing, we get the target $TDAG_{4}$, which can be used to
generate the English sentence ``John wished to walk.''

\section{Semantic Transfer}

As illustrated in the previous section, it is often the case that we
have to solve mismatches between source and target TDAGs in order to
obtain successful translations.
Syntactic/semantic transfer has been formulated by several
researchers\cite{shies90,zajac91} as a means of handling
situations in which fully
interlingual translation does not work. It is not enough, however,
to capture only
the equivalent relationship between source and target semantic
representations: this is merely a mapping among red nodes and arcs in TDAGs.
What is missing in the existing formulation is the provision of some
margin between {\it what is said} and {\it what is translated.} The semantic
transfer in our framework is defined as a set of successive operations on
TDAGs for creating a sequence of TDAGs $t_{0}$, $t_{1}$, $\ldots$,
$t_{k}$ such that $t_{0}$ is a source TDAG and $t_{k}$ is a target
TDAG that is a successful input to the sentence generator.

A powerful contextual processing and a domain knowledge base can be used
to infer additional facts and constraints, which correspond to the addition
of yellow nodes and arcs.
Default inheritance, proposed by Russell et al.\cite{russell92},
provides an efficient way of obtaining further information necessary
for translation, which corresponds to the addition of green nodes and
arcs.
A set of well-known heuristic rules, which we will describe later
in the ``Implementation'' Section, can also be used to add green
nodes and arcs.
To complete the model of semantic transfer, we have to
introduce a ``painter.'' A {\it painter} maps a red node to either a yellow
or a green node, a yellow node to a green node, and so on. It is used to
loosen the constraints imposed by the TDAGs. Every application of the
painter monotonically loses some information in a TDAG, and only a finite
number of applications of the painter are possible before the TDAG consists
entirely of green nodes and arcs except for a red root node. Note that the
painter never removes a node or an arc from a TDAG, it simply weakens
the constraints imposed by the nodes and arcs.

Formally, semantic transfer is defined as a sequence of the following
operations on TDAGs:
\begin{itemize}
\item Addition of a yellow node (and a yellow arc) to a given TDAG.
The node must be connected to a node in the TDAG by a yellow arc.
\item Addition of a yellow arc to a given TDAG. The arc must connect
two red or yellow nodes in the TDAG.
\item Addition of a green node (and a green arc) to a given TDAG.
The node must be connected to a node in the TDAG by the green arc.
\item Addition of a green arc to a given TDAG. The arc can connect
two nodes of any color in the TDAG.
\item Replacement of a red node (arc) with a yellow one, as long as the
well-formedness is preserved.
\item Replacement of a yellow node (arc) with a green one, as long as the
well-formedness is preserved.
\end{itemize}
The first two operations define the logical implications
(possibly with common sense or domain knowledge) of a given TDAG.
The next two operations define the defeasible (or heuristic) inference
from a given TDAG. The last two operations define the {\it painter}.
The definition of the painter specifies that it can
only gradually relax the constraints. That is, when a red or yellow
node (or arc) X
has other red or yellow nodes that are only connected through X,
X cannot be ``painted'' until each of the connected red and yellow nodes
is painted yellow or green to maintain the reachability through X.

In the sentence analysis phase, the first four operations can be
applied for obtaining
a source TDAG as a reasonable semantic interpretation
of a sentence. The application of these operations can be controlled
by ``weighted abduction''\cite{hobbs93}, default inheritance, and so on.
These operations can also be applied at semantic transfer for
augmenting the TDAG with a common sense knowledge of the target language.
On the other hand, these operations are not applied to a TDAG
in the generation phase, as we will explain in the next section. This is
because the lexicon and grammatical constraints are only applied to
determine whether red nodes and arcs are exactly derived.
If they are not exactly derived, we will
end up with either over- or under-generation beyond the permissible margin.
Semantic transfer is applied to a source TDAG as many times\footnote{The
iteration is bounded by the number of nodes and arcs in the TDAG, although
the number of possible sequences of operations could be exponential.}
as necessary until a successful generation is made. Recall the sample
sentence
in Figure~\ref{tdag}, where two {\it painter} calls were made to change
two red arcs in $TDAG_{1}$ into yellow ones in $TDAG_{2}$. These are
examples of the first substitution operation shown above.
An addition of a green node and a green arc, followed by an addition
of a green arc, was applied to $TDAG_{3}$ to obtain $TDAG_{4}$.
These additions are examples of the third and fourth addition operations.

\section{Sentence Generation Algorithm}

Before describing the generation algorithm, let us look at the
representation of lexicons and grammars for machine translation.
A {\it lexical rule}
is represented by a set of equations, which introduce red nodes and
arcs into a source TDAG.\footnote{For simplicity, we will only consider
{\it semantic equations} to form the TDAGs.} A {\it phrasal rule} is
similarly defined by a set of equations, which also introduce red nodes
and arcs for describing a syntactic head and its complements.

For example, if we use Shieber's
PATR-II\cite{shie86} notation, the lexical rule for {\mbox ``wished''} can
be represented as follows:

\begin{tabbing}
\renewcommand{\arraystretch}{.3}
V \= $\rightarrow$ \= wished \\
\> \tuple{V~cat} = v \\
\> \tuple{V~form} = past \\
\> \tuple{V~subj~cat} = np \\
\> \tuple{V~obj~cat} = v \\
\> \tuple{V~obj~form} = infinitival \\
\> \tuple{V~pred} = *WISH \\
\> \tuple{V~pred~agent} =~\tuple{V~subj~pred} \\
\> \tuple{V~pred~theme} =~\tuple{V~obj~pred} \\
\> \tuple{V~pred~theme~agent} =~\tuple{V~subj~pred} \\
\end{tabbing}

\begin{figure}
\centering{\epsfile{file=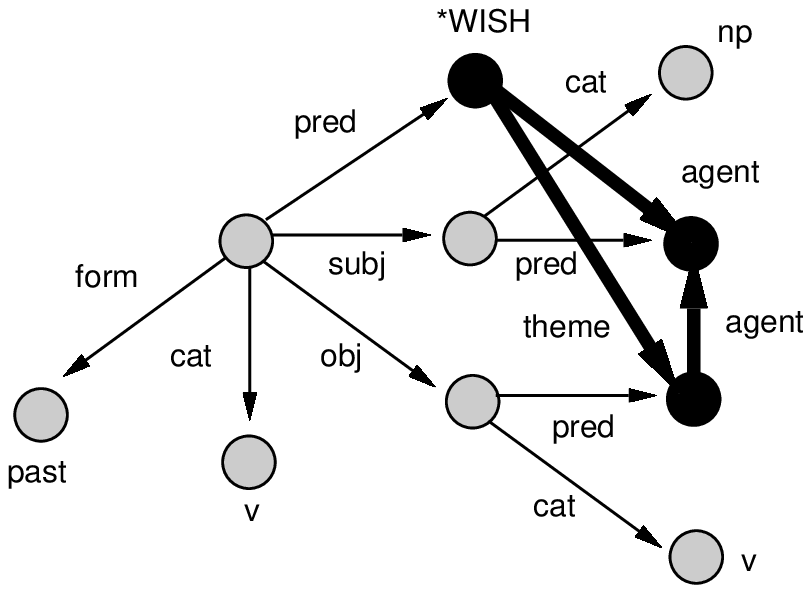,width=8cm,height=4.5cm}}
\caption{TDAG representation of the verb ``wished'' (embedded in the
entire feature structure)}
\label{lex}
\end{figure}

\begin{figure}
\centering{\epsfile{file=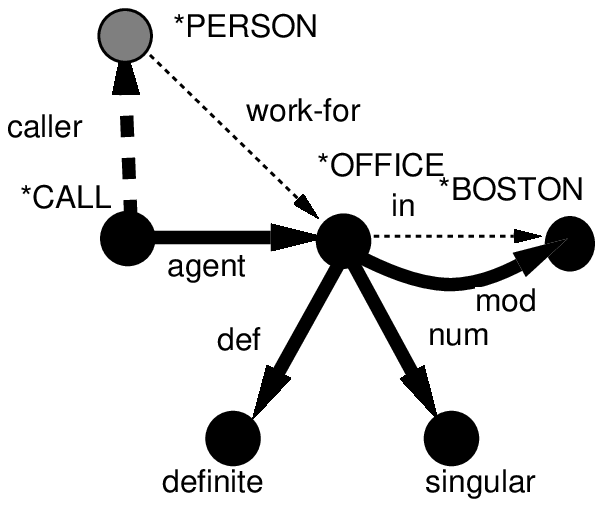,width=7cm,height=4cm}}
\caption{Source TDAG for the sentence ``The Boston Office called.''}
\label{hobbs}
\end{figure}

The last four equations are semantic equations.
Its TDAG representation is shown in Figure~\ref{lex}.
It would be more practical to further assume that such a lexical
rule is obtained
from a type inference system,\footnote{as in Shieber\cite{shie86},
Pollard and Sag\cite{polls87}, and Russell et al.\cite{russell92}}
which makes use of a syntactic class
hierarchy so that each lexical class can inherit general properties of
its superclasses. Similarly, semantic concepts such as *WISH and *WALK
should be separately defined in an ontological hierarchy together with
necessary domain knowledge (e.g., selectional constraints on case
fillers and {\it part-of} relationships. See KBMT-89\cite{kbmt89}.)
A unification grammar is used for both analysis and generation. Let us
assume that we have two unification grammars for English and Japanese.
Analyzing a sentence yields a source TDAG with red nodes and arcs.
Semantic interpretation resolves possible ambiguity and the resulting
TDAG may include all kinds of nodes and arcs. For example, the
sentence\footnote{in Hobbs et al.\cite{hobbs93}}
\begin{quote}
The Boston office called
\end{quote}
would give the source TDAG in Figure~\ref{hobbs}. By utilizing the
domain knowledge, the node labeled
*PERSON is introduced into the TDAG as a real caller of the action
*CALL, and two arcs representing {\it *PERSON work-for *OFFICE} and
{\it *OFFICE in *BOSTON} are abductively inferred.

Our generation algorithm is based on Wedekind's DAG traversal
algorithm\cite{wede88} for LFG.\footnote{It would be identical to
Wedekind's algorithm if an input TDAG consisted of only red nodes
and arcs.}
The algorithm runs with an input
TDAG by traversing the nodes and arcs that were derived from the
lexicon and grammar rules. The termination conditions are as follows:
\begin{itemize}
\item Every red node and arc in the TDAG was derived.
\item No new red node (arc) is to be introduced into the
TDAG if there is no corresponding node (arc) of any color in the TDAG.
That is,
the generator can change the color of a node (arc) to red,
but cannot add a new node (arc).
\item For each set of red paths (i.e., the sequence of red arcs)
that connects the same pair of nodes, the {\it reentrancy} was also
derived.
\end{itemize}
\begin{figure}
\centering{\epsfile{file=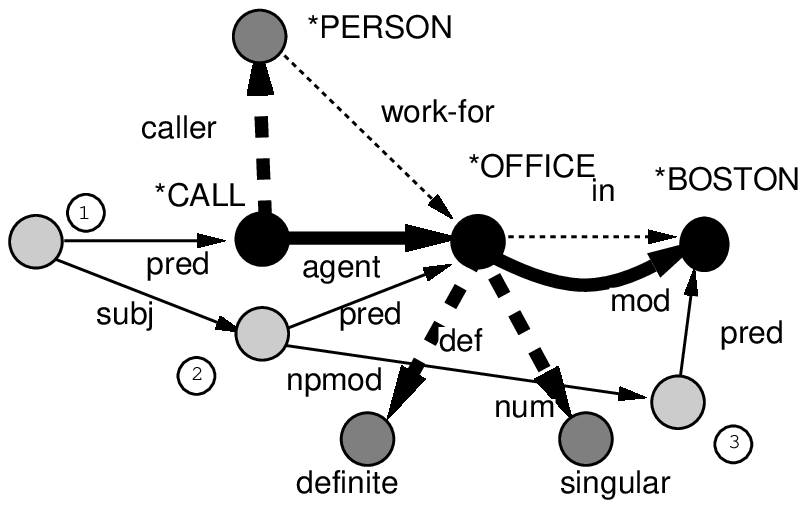,width=8cm,height=4.5cm}}
\caption{Target TDAG for the sentence ``The Boston Office called.''}
\label{hobbs2}
\end{figure}
These conditions are identical to those of Wedekind except that yellow
(or green) nodes and arcs may or may not be derived. For example,
the sentence ``The Boston Office called'' in Figure~\ref{hobbs}
can be translated into Japanese by the following sequence of
semantic transfer and sentence generation.
\begin{enumerate}
\item Apply the painter to change the yellow of the {\it definite} node
and the {\it def} arc to green.
\item Apply the painter to change the yellow of the {\it singular} node
and the {\it num} arc to green. The resulting TDAG is shown
in Figure~\ref{hobbs2}.
\item Run the sentence generator with an input feature structure,
which has a root and an arc {\it pred} connecting to the given TDAG.
(See the node marked ``1'' in Figure~\ref{hobbs2}.)
\item The generator applies a phrasal rule, say S $\rightarrow$ NP VP,
which derives the {\it subj} arc connecting to the subject NP (marked ``2''),
and the {\it agent} arc.
\item The generator applies a phrasal rule, say NP $\rightarrow$ MOD
NP,\footnote{There are several phrasal rules for deriving this LHS NP
in Japanese: (1) A noun-noun compound, (2) a noun, copula, and a noun,
and (3) a noun,
postpositional particle, and a noun. These three rules roughly
correspond to the forms (1) Boston Office, (2) office of Boston,
and (3) office in Boston. Inference of the {\it ``*OFFICE in *BOSTON''}
relation is easiest if rule (3) is used, but the noun-noun compound
is probably the best translation.}
which derives the {\it npmod} arc to the modifier of the NP
(marked ``3'') and the {\it mod} arc.
\item Lexical rules are applied and all the semantic nodes, *CALL,
*OFFICE, and *BOSTON are derived.
\end{enumerate}

\begin{figure}
{\small
\begin{verbatim}
;; run the generator with input f-structure
0>  *J-GG-START called with
((PRED "yobu") (CAT V) (VTYPE V-5DAN-B)
 (SUBCAT TRANS) (ASP-TYPE SHUNKAN)
 (:MOOD ((PRED "@dec")))
 (AUX ((PRED "@aux") (:TIME ((PRED "@past")))
   (:PASSIVE ((PRED "@minus")))))
 (SUBJ ((CAT N) (PRED "jimusho")
   (XADJUNCT ((XCOP "deno") (CAT N)
     (PRED "Boston"))))))
   ...
   3>  *J-GG-S called ;;<start> ->...-> <S>
    4>  *J-GG-XP called with ;;subj-filler
        ((CASE (*OR* "ha" "ga")) (CAT N)
         (NEG *UNDEFINED*) (PRED "jimusho")
         (XADJUNCT ((COP -) (CAT N)
            (PRED "Boston"))))
     5>  *J-GG-NP called ;;head NP of subj
          ...
          10<  *GG-N-ROOT returns ;;np mod
	       "Boston" ;;"Boston"
         9>  *J-GG-N called ;;head np
          10<  *GG-N-ROOT returns
	       "jimusho"
          ...
       7<  *9 (<SS> <NP>) returns ;;mod+NP
              "Boston deno jimusho"
        ...
     5<  *1 (<NP> <P>) returns ;;NP+case-marker
            "Boston deno jimusho ha"
    4<  *J-GG-XP returns "Boston deno jimusho ha"
    4>  *J-GG-S called with ;;VP part
     5>  *J-GG-VP called ;;stem +
      6>  *J-GG-V called ;;function word chains
      ((SUBJ *UNDEFINED*)
       (ADVADJUNCT *UNDEFINED*)
       (PPADJUNCT *UNDEFINED*)
       (:MOOD *UNDEFINED*)
       (AUX ((:TIME ((PRED "@past")))
        (:PASSIVE
         ((PRED (*OR* *UNDEFINED* "@minus"))))
        (PRED "@aux")))
       (CAT V) (TYPE FINAL) (ASP-TYPE SHUNKAN)
       (VTYPE V-5DAN-B) (SUBCAT TRANS)
       (PRED "yobu"))
       7>  *J-GG-RENTAI-PAST called ;;past-form
       ...
          14<  *GG-V-ROOT returns "yo" ;;stem
      ...
      6<  *J-GG-V returns "yobi mashita"
     5<  *J-GG-VP returns "yobi mashita"
    4<  *J-GG-S returns "yobi mashita"
   3<  *J-GG-S returns
       "Boston deno jimusho ha yobi mashita"
   ...
0<  *J-GG-START returns
    "Boston deno jimusho ha yobi mashita"
\end{verbatim}}
\caption{Sentence generation from the TDAG for ``The Boston Office called.''}
\label{gen}
\end{figure}

The annotated sample run of the sentence
generator is shown in Figure~\ref{gen}. The input TDAG in the sample run
is embedded in the input feature structure as a set of PRED values,
but the semantic arcs are not shown in the figure. The input feature
structure has syntactic features that were specified in the lexical
rules. The feature value *UNDEFINED* is used to show that the node has
been traversed by the generator.

The basic property of the generation algorithm is as follows:
\begin{quote}
Let $t$ be a given TDAG, $t_{min}$ be the connected subgraph including all
the red nodes and arcs in $t$, and $t_{max}$ be the connected subgraph
of $t$ obtained by changing all the colors of the nodes and arcs to red.
Then, any successful generation with the derived TDAG $t_{g}$ satisfies
the condition that $t_{min}$ subsumes $t_{g}$, and $t_{g}$
subsumes $t_{max}$.
\end{quote}
The proof is immediately obtained from the definition of successful
generation and the fact that the generator never introduces a new
node or a new arc into an input TDAG.
The TDAGs can also be employed by the semantic head-driven generation
algorithm\cite{shiep90} while retaining the above property. Semantic
monotonicity always holds for a TDAG, since red nodes must be connected.
It has been shown by Takeda\cite{take9312} that semantically non-monotonic
representations can also be handled by introducing a {\it functional}
semantic class.

\section{Implementation}

We have been developing a prototype English-to-Japanese MT system, called
Shalt2\cite{takeu9207}, with a lexicon for a computer-manual domain
including about 24,000 lexemes each for English and Japanese, and a
general lexicon including about 50,000 English words and their translations.
A sample set of
736 sentences was collected from the ``IBM AS/400 Getting Started'' manual,
and was tested with the above semantic transfer and generation
algorithm.\footnote{We used McCord's English parser based on his
English Slot Grammar\cite{mccord90}, which covered more than 93\% of the
sentences.}
The result of the syntactic analysis by the English parser is mapped
to a TDAG using a set of semantic equations\footnote{We call such a
set of semantic equations
{\it mapping rules} (see Shalt2\cite{take9307} or KBMT-89\cite{kbmt89}).}
obtained from the lexicons. We have a very shallow knowledge base
for the computer domain, and no logical inference system was used
to derive further constraints from the given source sentences.
The Japanese grammar is similar to the one used in KBMT-89, which is
written in pseudo-unification\cite{tomik87} equations, but we have added
several new types of equation for handling coordinated structures.
The Japanese grammar can generate sentences from all the successful
TDAGs for the sample English sentences.

It turned out that there were a few collections of semantic
transfer sequences which contributed very strongly to the successful
generation. These sequences include
\begin{itemize}
\item Painting the functional control arcs in yellow.
\item Painting the gaps of relative clauses in yellow.
\item Painting the number and definiteness features in yellow.
\item Painting the passivization feature in green.\footnote{We decided
to include the passivization feature in the semantic representation
in order to determine the proper word ordering in Japanese.}
\end{itemize}
Other kinds of semantic transfer are rather idiosyncratic, and are
usually triggered by
a particular lexical rule. Some of the sample sentences used for
the translations are as follows:\footnote{Japanese translation reflects
the errors made in English analysis. For example, the auxiliary
verb ``could'' is misinterpreted in the last sample sentence.}
\begin{verbatim}
Make sure you are using the proper edition
for the level of the product.

YUUZAA HA  SEIHIN   NO  REBERU NI
user +subj product +pos level +for
TEKISETSUNA HAN      WO  SIYOUSHITE  IRU
proper      edition +obj use        +prog
KOTO  WO  TASHIKAMETE KUDASAI
+nom +obj confirm    +imp

Publications are not stocked at the address
given below.

SIRYOU       HA   IKA        DE  TEIKYOUSURU
publication +subj following +loc provide
ADORESU  NI  SUTOKKU  SARE     MASEN
address +loc stock   +passive +neg

This publication could contain technical
inaccuracies or typographical errors.

KONO SIRYOU       HA   GIJYUTSUTEKINA
this publication +subj technical
HUSEIKAKUSA ARUIHA INSATSUJYOUNO ERAA   WO
inaccuracy  or     typographical error +obj
FUKU     ME       MASHITA
contain +ability +past
\end{verbatim}

The overall accuracy of the translated sentences
was about 63\%. The main reason for translation
errors was the occurrence of errors in lexical and structural
disambiguation by the syntactic/semantic analyzer.
We found that the accuracy of semantic transfer and sentence
generation was practically acceptable.

Though there were few serious errors, some occurred
when a source TDAG had to be completely ``paraphrased''
into a different TDAG. For example, the sentence
\begin{verbatim}
Let's get started.
\end{verbatim}
was very hard to translate into a natural Japanese sentence.
Therefore, a TDAG had to be paraphrased into a totally different TDAG,
which is another important role of semantic transfer.
Other serious errors were related to the
ordering of constituents in the TDAG. It might be generally
acceptable to assume that the ordering of nodes in a DAG is immaterial.
However, the different ordering of adjuncts sometimes resulted
in a misleading translation, as did the ordering of members in
a coordinated structure. These subtle issues have to be taken into
account in the framework of semantic transfer and sentence generation.

\section{Conclusions}
In this paper, we have introduced tricolor DAGs to represent various
degrees of constraint, and defined the notions of semantic transfer
and sentence generation as operations on TDAGs. This approach
proved to be so practical that nearly all of the source sentences
that were correctly parsed were translated into readily acceptable
sentences.
Without semantic transfer, the translated sentences would include
greater numbers of incorrectly selected words, or in some cases
the generator would simply fail\footnote{The Essential Arguments
Algorithm\cite{martin92} might be an alternative method for finding
a successful generation path.}

Extension of TDAGs for disjunctive information and a set of feature
structures must be fully incorporated into the framework. Currently only
a limited range of the cases are implemented. Optimal control of semantic
transfer is still unknown.
Integration of the constraint-based formalism,
defeasible reasoning, and practical heuristic rules are also important
for achieving high-quality translation. The ability to process and
represent various levels of knowledge in TDAGs by using
a uniform architecture is desirable, but there appears to be
some efficient procedural knowledge that is very
hard to represent declaratively. For example, the
negative determiner ``no'' modifying a noun phrase in English has to be
procedurally transferred into the negation of the verb governing the
noun phrase in Japanese. Translation of ``any'', ``yet'', ``only'',
and so on involves similar problems.

While TDAGs reflect three discrete types of constraints, it is
possible to generalize the types into continuous, numeric values
such as {\it potential energy}\cite{hasida92}. This approach will
provide a considerably more flexible margin that defines a set of
permissible translations, but it is not clear whether we can successfully
define a numeric value for each lexical rule in order to obtain acceptable
translations.

\section{Acknowledgments}

The idea of the tricolor DAGs grew from discussions
with Shiho Ogino on the design and implementation of the
sentence generator. I would also like to thank the members
of the NL group -- Naohiko Uramoto, Tetsuya Nasukawa, Hiroshi Maruyama,
Hiroshi Nomiyama, Hideo Watanabe, Masayuki Morohashi, and Taijiro
Tsutsumi -- for stimulating comments and discussions that
directly and indirectly contributed to shaping the paper.
Michael McDonald, who has always been the person I turn to
for proofreading, helped me write the final version.


\begin{thebibliography}{10}

\bibitem{dymetman91}
M.~Dymetman.
\newblock {``Inherently Reversible Grammars, Logic Programming and
  Computability''}.
\newblock In {\em Proc. of ACL Workshop on Reversible Grammar in Natural
  Language Processing}, pages 20--30, Berkeley, California, June 1991.

\bibitem{emele92}
M.~Emele, U.~Heid, S.~Momma, and R.~Zajac.
\newblock {``Interactions between Linguistic Constraints: Procedural vs.
  Declarative Approaches''}.
\newblock {\em Machine Translation}, 7(1-2):61--98, 1992.

\bibitem{hasida91}
K.~Hasida.
\newblock {``Common Heuristics for Parsing, Generation, and Whatever,
  $\ldots$''}.
\newblock In {\em Proc. of a Workshop on Reversible Grammar in Natural
  Language Processing}, pages 81--90, June 1991.

\bibitem{hasida92}
K.~Hasida.
\newblock {``Dynamics of Symbol Systems - An Integrated Architecture of
  Cognition -''}.
\newblock In {\em Proc. of International Conference on Fifth Generation
  Computer Systems 1992}, pages 1141--1148, June 1992.

\bibitem{hobbs93}
J.~R. Hobbs, M.~E. Stickel, D.~E. Appelt, and P.~Martin.
\newblock {``Interpretation as abduction''}.
\newblock {\em Artificial Intelligence}, 63:69--142, 1993.

\bibitem{kaplb82}
R.~Kaplan and J.~Bresnan.
\newblock {``Lexical-Functional Grammar: A Formal System for Generalized
  Grammatical Representation''}.
\newblock In J.~Bresnan, editor, {\em {``Mental Representation of Grammatical
  Relations''}}, pages 173--281. MIT Press, Cambridge, Mass., 1982.

\bibitem{kaspr8606}
R.~Kasper and W.~C. Rounds.
\newblock {``A Logical Semantics for Feature Structures''}.
\newblock In {\em Proc. of the 24th Annual Meeting of the Association for
  Computational Linguistics}, Columbia University, New York, NY, June 1986.

\bibitem{kbmt89}
KBMT89.
\newblock {``Special Issue on Knowlege-based Machine Translation I and II''}.
\newblock {\em Machine Translation}, 4(2-3), March-June 1989.

\bibitem{martin92}
M.~Martinovic and T.~Strzalkowski.
\newblock {``Comparing Two Grammar-Based Generation Algorithms: A Case
  Study''}.
\newblock In {\em Proc. of the 30th Annual Meeting of ACL}, pages 81--88,
  June 1992.

\bibitem{mccord90}
M.~McCord.
\newblock {\em {``Slot Grammar: A System for Simpler Construction of
  Practical Natural Language Grammars (Ed:Studer,R.)''}}, pages 118--145.
\newblock Springer-Verlag, 1990.

\bibitem{murata93}
M.~Murata and M.~Nagao.
\newblock {``Determination of Referential Property and Number of Nouns in
  Japanese Sentences for Machine Translation into English''}.
\newblock In {\em Proc. of the 5th International Conference on Theoretical
  and Methodological Issues in Machine Translation}, pages 218--225, Kyoto,
  Japan, July 1993.

\bibitem{kant}
E.~H. {Nyberg, 3rd} and T.~Mitamura.
\newblock {``The KANT System: Fast, Accurate, High-Quality Translation in
  Practical Domains''}.
\newblock In {\em Proc. of the 14th International Conference on Computational
  Linguistics}, pages 1069--1073, July 1992.

\bibitem{polls87}
C.~Pollard and I.~A. Sag.
\newblock {\em {``An Information-Based Syntax and Semantics, Vol.1
  Fundamentals''}}.
\newblock CSLI Lecture Notes, Number 13, 1987.

\bibitem{russell92}
G.~Russell, A.~Ballim, J.~Carroll, and S.~Warwick-Armstrong.
\newblock {``A Practical Approach to Multiple Default Inheritance for
  Unification-Based Lexicons''}.
\newblock {\em Computational Linguistics}, 18(3):311--337, Sept. 1992.

\bibitem{shie86}
S.~M. Shieber.
\newblock {\em {``An Introduction to Unification-Based Approaches to
  Grammar''}}.
\newblock CSLI Lecture Notes, Number 4, Stanford, CA, 1986.

\bibitem{shie88}
S.~M. Shieber.
\newblock {``A Uniform Architecture for Parsing and Generation''}.
\newblock In {\em Proc. of the 12th International Conference on Computational
  Linguistics}, pages 614--619, August 1988.

\bibitem{shiep90}
S.~M. Shieber, F.~C.~N. Pereira, G.~van Noord, and R.~C. Moore.
\newblock {``Semantic-Head-Driven Generation''}.
\newblock {\em Computational Linguistics}, 16(1):30--42, March 1990.

\bibitem{shies90}
S.~M. Shieber and Y.~Schabes.
\newblock {``Synchronous Tree-Adjoining Grammars''}.
\newblock In {\em Proc. of the 13th International Conference on Computational
  Linguistics}, pages 253--258, August 1990.

\bibitem{smol88}
G.~Smolka.
\newblock {``A Feature Logic with Subsorts''}.
\newblock Technical Report LILOG-REPORT 33, IBM Deutschland GmbH, Stuttgart,
  West Germany, May 1988.

\bibitem{take9307}
K.~Takeda.
\newblock {``An Object-Oriented Implementation of Machine Translation
  Systems''}.
\newblock In {\em Proc. of the 5th International Conference on Theoretical
  and Methodological Issues in Machine Translation}, pages 154--167,
  July 1993.

\bibitem{take9312}
K.~Takeda.
\newblock {``Sentence Generation from Partially Constrained Feature
  Structures''}.
\newblock In {\em Proc. of the Natural Language Processing Pacific Rim
  Symposium}, pages 7--16, Dec. 1993.

\bibitem{takeu9207}
K.~Takeda, N.~Uramoto, T.~Nasukawa, and T.~Tsutsumi.
\newblock {``Shalt2 - A Symmetric Machine Translation System with Conceptual
  Transfer''}.
\newblock In {\em Proc. of the 14th International Conference on Computational
  Linguistics}, pages 1034--1038, July 1992.

\bibitem{tomik87}
M.~Tomita and K.~Knight.
\newblock {``Pseudo Unification and Full Unification''}.
\newblock Technical Report CMU-CMT-88-MEMO, Center for Machine Translation,
  Carnegie Mellon University, November 1987.

\bibitem{uchida88}
H.~Uchida.
\newblock {``ATLAS II: A Machine Translation System Using Conceptual
  Structure as an Interlingua''}.
\newblock In {\em Proc. of 2nd Intl. Conf. on Theoretical and Methodological
  Issues in Machine Translation of Natural Languages}, pages 150--160, June
  1988.

\bibitem{wede88}
J.~Wedekind.
\newblock {``Generation as Structure Driven Derivation''}.
\newblock In {\em Proc. of the 12th International Conference on Computational
  Liguistics}, pages 732--737, August 1988.

\bibitem{yasuhara93}
H.~Yasuhara.
\newblock {``Conceptual Transfer in an Interlingua Method and Example Based
  MT''}.
\newblock In {\em Proc. of the Natural Language Processing Pacific Rim
  Symposium '93}, pages 376--379, Fukuoka, Japan, Dec. 1993.

\bibitem{zajac91}
R.~Zajac.
\newblock {``A Uniform Architecture for Parsing, Generation and Transfer''}.
\newblock In {\em Proc. of a Workshop on Reversible Grammar in Natural
  Language Processing}, pages 71--80, June 1991.

\end{thebibliography}
\end{document}